\documentclass[smallcondensed]{svjour3}
\journalname{Machine Learning}

\usepackage{graphicx}
\usepackage{latexsym}
\usepackage{amsmath,amsfonts}
\usepackage{color}
\usepackage[ruled,vlined]{algorithm2e}
\usepackage{subcaption}
\usepackage[colorlinks=True]{hyperref}

\def\urltilda{\kern -.15em\lower .7ex\hbox{\~{}}\kern .04em}

\newtheorem{myexample}{Example}

\SetCommentSty{mycommfont}

\begin{document}

\pagestyle{headings}


\title{Incorporating Symbolic Domain Knowledge into Graph Neural Networks}

\author{Tirtharaj Dash \and 
        Ashwin Srinivasan \and
        Lovekesh Vig
}


\institute{T. Dash \and A. Srinivasan \at
            APP Centre for Artificial Intelligence Research \\
            Department of Computer Science \& Information Systems \\
            BITS Pilani, K.K. Birla Goa Campus, Goa\\
            \email{\{tirtharaj,~ashwin\}@goa.bits-pilani.ac.in}
            \and
            L. Vig \at
            TCS Innovation Labs, New Delhi
}

\maketitle

\begin{abstract}
Our interest is in scientific problems with the following
characteristics: (1) Data are naturally represented as graphs;
(2) The amount of data available is typically small; and
(3) There is significant
domain-knowledge, usually expressed in some symbolic form (rules, taxonomies, constraints
and the like). These kinds of problems have been addressed effectively in the past by
symbolic machine learning methods like Inductive Logic Programming (ILP), by virtue
of 2 important characteristics: (a) The use of a representation language that easily
captures the relation encoded in graph-structured data, and (b) The inclusion
of prior information encoded as domain-specific relations, that can alleviate problems of
data scarcity, and construct new relations. Recent advances have seen the emergence
of deep neural networks specifically developed for graph-structured data (Graph-based
Neural Networks, or GNNs). While GNNs have been shown to
be able to handle graph-structured data, less has been
done to investigate the inclusion of domain-knowledge. Here we investigate this aspect of GNNs empirically
by employing an operation we term  {\em vertex-enrichment\/}
and denote the corresponding GNNs as $VEGNN$s.
Using over 70 real-world datasets
and substantial amounts of symbolic domain-knowledge, we examine the result of
vertex-enrichment across 5 different variants of GNNs. Our results provide
support for the following:
(a) Inclusion of domain-knowledge by vertex-enrichment can significantly
    improve the performance of a GNN.
    That is, the performance of $VEGNN$s is significantly better than $GNN$s across all GNN variants;
(b) The
    inclusion of domain-specific relations constructed using ILP improves the performance of $VEGNNs$, across all
    GNN variants.
Taken together, the results provide evidence that it is possible to
incorporate symbolic domain knowledge into a GNN, and
that ILP can play an important role in providing high-level
relationships that are not easily discovered by a GNN.
\end{abstract}

\section{Introduction}
Industrialising scientific discovery, in the manner demonstrated by the
Robot Scientist Project~\cite{king2004functional} uses machine learning programs as
scientific assistants. At the very least, this would appear to require machine learning methods that are able to (a) cope with data that have some inherent
structure, in the form of entities and relations;
and (b) construct good predictive models
by effectively drawing on any existing scientific knowledge thought to
be relevant. A really useful
assistant would have to do more. A wish-list would
include identifying the best explanation for a prediction based on
what is known; suggesting hidden variables or mechanisms which could
improve the prediction; and proposing experiments to test the hypotheses.
The Robot Scientist Project showed ways to achieve each of these
in some measure with Inductive Logic Programming (ILP). Recent rapid gains in
neural-network technology suggest that deep networks could form the
basis of extremely powerful predictive models, which is clearly relevant to the construction of an effective scientific assistant.
Here we investigate the performance of state-of-the-art
deep networks specifically designed to analyse graph-structured
data. A substantial number of applications addressed by ILP belong
to this category of data (see, for example,~\cite{king1996structure,srinivasan1999feature,faruquie2012topic}). There
are at least three good reasons to investigate if graph neural networks,
or GNNs, are able to incorporate domain-knowledge. First, studies with
ILP have repeatedly shown that inclusion of domain-knowledge can make  substantial
difference to predictive performance. Furthermore, a recent report on
Artificial Intelligence (AI) for Science identifies incorporating domain-knowledge in AI as
one of the three Grand Challenges facing 
the application of AI~\cite{stevens2020ai}.
Deep learning methods based on neural networks have not
focused on this, relying instead on their internal computational machinery to
construct higher-level concepts automatically from
the raw data. The ILP experience suggests
otherwise, and we would like to know if this applies to GNNs.
Second, symbolic encodings of domain knowledge are both natural and flexible
ways of encoding prior knowledge. ILP systems implemented as logic programs have been the pre-eminent
form of machine learning for using such knowledge. Despite extremely efficient
implementations of logic programming, the significant
world-wide effort into the development of deep learning tools that have resulted in
highly efficient implementations that exploit the
processing capabilities of graphics processing units (GPUs). A GNN capable of including
symbolic domain knowledge could provide an efficient way of constructing predictive
models. Thirdly, to the best of our knowledge, GNN applications to date have been restricted to
simple node-and-edge features, and have not attempted to encode
any significant domain-knowledge. The real-world problems we examine in this paper
have very extensive amounts of domain information, resulting from many years of academic and industrial effort into the use of ILP.

In this paper, we restrict the investigation to the problem of
prediction, which we see as a necessary first step in the
development of automated scientific assistants. Facilities for
explanations and experiment-proposal using GNN models are
conceptually harder, are deferred to future work. We assess
the use of domain-knowledge using a sample of over 70 datasets,
containing over 200,000 data instances. The datasets refer to
problems in a broad category known as structure-activity prediction.
Each data instance is, therefore, a molecule, which is naturally
represented as a graph.
\footnote{In fact, GNNs were originally tested with molecular datasets~\cite{baskin1997neural}.} 
For this class of problems, we
now have a sufficiently large body of domain-knowledge encoded in human-understandable symbolic relations. This allows us to perform a
case-study on the inclusion of symbolic relations by GNNs. The
principal contributions of the paper are as follows:

\begin{itemize}
    \item To the field of graph neural networks, the paper
        presents a large-scale empirical study using real-world datasets on the inclusion of domain-knowledge. To the best of our knowledge, the number of graphs used and the number of relations encoding domain-knowledge are the most extensive to date.
    \item To the field of inductive logic programming, the paper demonstrates
        a continuing case for the usefulness  of ILP on relational learning
        tasks, despite the development of very efficient deep neural
        networks specifically designed for a specific form of relational
        data. 
    \item To the field of neuro-symbolic modelling, the technique of
        vertex-enrichment described in the paper provides a simple but
        effective way of incorporating symbolic relations into graph-based
        neural networks.
\end{itemize}

The rest of the paper is organised as follows. In Section \ref{sec:gnn}, we describe vertex-enrichment in graphs and a set of practical considerations arising from
the developed algorithms. In Section \ref{sec:expt},
we describe our aims, data and background knowledge, 
the specifics of the methodology, and the
obtained results. Section \ref{sec:relwork}
lists related works, and Section \ref{sec:concl} concludes the paper.

\section{Graph Neural Networks (GNNs)}
\label{sec:gnn}

GNNs are primarily developed for learning
from data represented as graphs. For completeness,
we include some basic definitions first.

\begin{definition}[Graphs]
A graph $G$ is a pair $(V,E)$ where $V$ is a set
of vertices, $E$ is a set of edges and a subset of $V \times V$. A graph is 
said to be undirected if for every $(v_i,v_j) \in E$, $(v_j,v_i)$ in $E$.
\end{definition}

We will be concerned in this paper with undirected graphs.
We note that for such graphs, $E$ can be represented more compactly
as a set consisting of 1- or 2-element subsets of $V$. We will return to this
later, as we extend the consideration to hypergraphs. For molecular graphs, of the kind considered here, self-loops do not occur.



\begin{myexample}
\label{ex:molgraph}
{\bf Molecules as graphs.}
A benzene ring (shown below) can be represented as a graph, in which
vertices correspond to atoms and edges correspond to bonds~\cite{mcnaught1997compendium}.
\begin{center}
\includegraphics[width=0.4\textwidth]{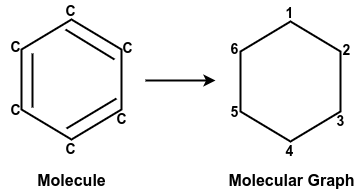}
\end{center}
The graph-representation of the molecule on the left is:
\[
\begin{split}
    (\{1,2,3,4,5,6\},\{(1,2),(2,1),(2,3),(3,2),(3,4),
    (4,3),(4,5),(5,4),(5,6),(6,5),\\
    (6,1),(1,6)\})
\end{split}
\]
\end{myexample}

\noindent
We will need the concept of the {\em neighbourhood} of a vertex
in an undirected graph:\footnote{Henceforth, by ``graph'' we will mean an undirected graph.}

\begin{definition}[Neighbourhood]
Given a graph $G = (V,E)$, $\sigma$ is a 
neighbourhood function from $V$ to $2^V$.
\end{definition}

\begin{myexample}
One obvious definition of $\sigma$ for an undirected graph $(V,E)$
is $\sigma(v) = \{v_i: v_i \in V, (v,v_i) \in E\}$.
For the graph in Example~\ref{ex:molgraph}, 
$\sigma(1)=\{2,6\}, \sigma(2)=\{1,3\}$.
\end{myexample}
For the GNNs in this paper, we will need labelled undirected graphs.

\begin{definition}[Graph Labellings]
Let ${\cal V}$ be a set of vertex labels
and ${\cal E}$ be a set of edge labels.
Then a vertex-labelling of a graph $G = (V,E)$ is
a function $\psi: V \rightarrow 2^{\cal V}$ and
an edge-labelling is a function $\epsilon:E \rightarrow 2^{\cal E}$.\footnote{
We do not commit here to any specific data structure that
should be used to implement the label set. This could be,
for example, a Boolean-valued array
of size $|{\cal V}|$.
}
\end{definition}

\begin{myexample}
The vertex labels of the graph given in Example~\ref{ex:molgraph}
can be the atom-types (Carbon, C), and edge labels can be the bond-types (single bond: 1, double bond: 2). The label for the vertex 1 is $\psi(1)=\cdots=\psi(6)=\{C\}$. The labelling for the edges are $\epsilon((1,2))=\epsilon((2,1))= \{2\}$, $\epsilon((2,3))=\epsilon((3,2))=\{1\}$ and so on.

\end{myexample}
Although not evident in this example, vertex- and edge-labels can have more than one element (hence the
mapping to $2^{\cal V}$ and $2^{\cal E})$. This will
be necessary later.

We will use the term {\em graph} interchangeably to denote
the tuple $(V,E)$ or the tuple
$(V,E,\sigma,\psi,\epsilon)$.
We are interested here in classifying graphs. That is,
given a set of class labels ${\cal Y}$, we want
to construct a function that maps a graph of the form
$(V,E,\sigma,\psi,\epsilon)$ to ${\cal Y}$. A GNN is one such
function that employs 2 higher-order functions.

\begin{definition}[Relabel]
Given a graph
$(V,E,\sigma,\psi,\epsilon)$.
Let $Relabel$ be a function that returns a graph $(V,E,\sigma,\psi',\epsilon')$, where
the functions $\psi'$ and $\epsilon'$
may be different to $\psi$ and $\epsilon$.
\end{definition}

\noindent
A vectorisation function is used to map
a graph as a real-valued vector.

\begin{definition}[Vectorise]
Let ${\cal G}$ denote
the set of graph-tuples of the form
$(V,E,\sigma,\psi,\epsilon)$.
A vectorisation of the graph-tuple is
the result of applying a function
$Vec: {\cal G} \rightarrow \Re^d$ ($d \geq 1$). 
\end{definition}

A GNN is the composition of these functions, and
some prediction function as implemented by a neural network.

\begin{definition}[GNN]
\label{def:gnn}
Let $NN: \Re^d \rightarrow {\cal Y}$ denote
a neural network that maps a real-valued vector to a set of
class labels. Given a $G$ = $(V,E,\sigma,\psi,\epsilon)$,
$GNN(G) = NN(Vec(Relabel(G)))$.
\end{definition}

\noindent
Variations of GNNs result from changing the definitions of $NN, Vec$ and $Relabel$. Many different
definitions of the $Relabel$ function have been proposed recently. We defer
the specific details of the GNN variants used here to Section \ref{sec:gnn_prac}.

\subsection{Encoding $n$-ary Relations}
\label{sec:embed}

GNNs, as we have described them so far, deal with
node- and edge-labels in an undirected graph, in which edges are
sets of vertex-pairs. That is, the edges represent a symmetric binary
relation. However, for many
real-world problems---including the ones considered in
this paper---we have access to domain-knowledge which relate
more than just pairs of vertices. For example, if a molecule
is represented as a graph (with atoms as vertices, and an
edge denoting a bond between a pair of vertices), then
a benzene-ring is a relation amongst 6 distinct vertices,
with some specific constraints on the vertices and edges. 
Here, we will consider domain-knowledge to be a set of relations,
each of which can be expressed as a hypergraph.

\begin{definition}[Hypergraphs]
A hypergraph $H$ is the pair $(V,E')$,
where $V$ is a set of vertices
and $E'$ is a non-empty subset of $2^V$. Each
element of $E'$ is called a hyperedge.
\end{definition}

\begin{myexample} 
\label{ex:hypgraph}
A hypergraph of the molecular graph given in
Example~\ref{ex:molgraph} can be
$H = (\{1,2,3,4,5,6\},\{\{1,2\},\{3,4,5,6\},\{2,4,5\},\{1,2,3,4,5,6\}\})$.
\end{myexample}

\noindent
We note that since hyperedges are sets, there
is no distinction between permutations of vertices
in a hyperedge. So, as defined here, we will
take hyperedges as being undirected. Hypergraph
labellings can be defined similarly as before, using
a pair of functions for vertex- and edge-labels. 
We will reuse the notation $\psi$ and $\epsilon$ for these functions,
with annotations to clarify what is meant. The neighboorhood
relation $\sigma$ is left unspecified here (one
obvious definition is
$\sigma(v_i) = \{v_j: h \in E', \{v_i,v_j\} \subseteq h\}$). 
In this paper, we are interested in $n$-ary
relations that can be expressed as hypergraphs.

\begin{definition}[$n$-ary Relation as a Labelled Hypergraph]
A $n$-ary relation $R$ defined over vertices
of a graph $G = (V,E)$
is a hypergraph $H = (V,E')$, 
and every hyperedge $h \in E'$ has $n$ elements from $V$.
We will denote this as $R(G) = H$. Let $\psi_G$
denote a vertex-labelling over $G$ and $R/n$ denote
the predicate-symbol for $R$. With
some abuse of notation, the vertex-labelling function
for $R(G) = H = (V,E')$ is as follows:

\begin{align*}
\psi_H(v) = \left\{ \begin{array}{cc} 
                \psi_G(v) \cup \{R/n\} & \hspace{5mm} \mathrm{if}~ \exists h \in E' s.t.~ v \in h \\
                \emptyset & \hspace{5mm} {\mathrm{otherwise}} \\
                \end{array} \right.
\end{align*}

and the hyperedge-labelling function is:

\[
    \epsilon_H(h) ~=~ \{R/n\}~~~(h \in E')
\]
\end{definition}

\noindent
That is, the vertex-labelling of a vertex $v$ in the hypergraph $H$ is a set containing the existing vertex-label
of $v$ in $G$ augmented by the predicate-symbol $R/n$
vertex-label. 
\begin{myexample}
Consider a relation for a Benzene ring:
\begin{align*}
    Benzene(a_1,a_2,a_3,a_4,a_5,a_6) &\leftarrow \\ 
    & Cycle(a_1,a_2,a_3,a_4,a_5,a_6) ~\wedge\\
    & Aromatic(a_1,a_2,a_3,a_4,a_5,a_6).
\end{align*}
One possible vertex-labelling is:
\[
    \psi_H(1)=\cdots=\psi_H(6)=\{C,Benzene/6\}
\]
(here, $C$ denotes ``carbon''). A hyperedge-labelling may contain:
\[
    \epsilon_H(\{1,2,3,4,5,6\}) = \{Benzene/6\}
\]
\end{myexample}

\noindent
The extension to multiple relations, not all of the same arity,
is straightforward.

\begin{definition}[Multiple Relations as a Labelled Hypergraph]
\label{def:multirelasH}
Let $R_1,\ldots, R_k$ be relations defined on vertices
of a graph $G = (V,E)$, s.t. $R_i(G) = (V,{E_i}')$.
Then $\bigcup R_i(G)$ is the hypergraph $H = (V,{E}')$
where $E' = \bigcup {E_i}'$.
The corresponding labelling functions are:

\[
    \psi_H(v) = \bigcup \psi_{H_i}(v) 
\]

\noindent
and
\[
    \epsilon_H(v) = \bigcup \epsilon_{H_i}(v) 
\]
\end{definition}

\begin{myexample}\label{ex:molgraph2}
In the molecular graph given below, there are two relations: $Benzene/6$ and $Pyrrole/5$.

\begin{center}
\includegraphics[width=0.2\textwidth]{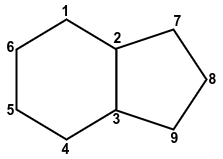}
\end{center}

One possible vertex-labelling for this graph is:
\begin{align*}
    \psi_H(1)=\psi_H(4)=\psi_H(5)=\psi_H(6)=\{C, Benzene/6\} \\
    \psi_H(8)=\psi_H(9)=\{C, Pyrrole/5\} \\
    \psi_H(7)=\{N,Pyrrole/5\}\\
    \psi_H(2)=\psi_H(3)=\{C,Benzene/6,Pyrrole/5\}
\end{align*}

and a hyperedge-labelling is:
\begin{align*}
    \epsilon_H(\{1,2,3,4,5,6\}) = \{Benzene/6\} \\
    \epsilon_H(\{2,7,8,9,3\}) = \{Pyrrole/5\}
\end{align*}
\end{myexample}

\noindent
In principle, provided we are able to define
a neighbourhood function $\sigma$ for hypergraphs,
the definition of GNNs in Defn.~\ref{def:gnn} does not change.
We would however like to use one of the standard GNN implementations
described in the previous section, which restricts graphs with
2-vertex edges, and edge-labels to singleton sets. With some loss of information, we extract a suitable graph from a hypergraph.

\begin{definition}
{\bf (Vertex-Enriched Graphs)}
Let $G = (V,E)$ be a graph, with neighbourhood
function $\sigma$, vertex-labelling function
$\psi$, and edge-labelling function $\epsilon$.
Here, $E$ is a subset of $V \times V$.
Let $\mathcal{R} = \{R_1,\ldots,R_k\}$ be a set of relations defined on $G$,
and $\bigcup R_i(G)$ be the hypergraph $H$ = $(V,E')$
with vertex-labelling function $\psi'$
as in Defn. \ref{def:multirelasH}. Then
$G'$ = $(V,E,\sigma,\psi',\epsilon)$ is called
a vertex-enriched form of $G$ = $(V,E,\sigma,\psi,\epsilon)$.
We denote this by $VE(G, \mathcal{R}) = G'$.
\end{definition}

\begin{myexample}
The molecular graph $G$ for Example~\ref{ex:molgraph2} is
\[
\begin{split}
    G=(\{1,2,3,4,5,6,7,8,9\},\{(1,2),(2,1),\cdots,(1,6),(6,1),(2,7),(7,2),\cdots,\\
    (9,3)(3,9)\})
\end{split}
\]
A vertex-labelling of $G$ is:
\begin{align*}
    \psi(1)=\cdots=\psi(6)=\psi(8)=\psi(9)=\{C\}\\
    \psi(7)=\{N\}
\end{align*}
The vertex-labelling of the vertex-enriched graph $G'$, after the
inclusion of the relations in Example~\ref{ex:molgraph2} is:

\begin{align*}
    \psi'(1)=\psi'(4)=\psi'(5)=\psi'(6)=\{C, Benzene/6\} \\
    \psi'(8)=\psi'(9)=\{C, Pyrrole/5\} \\
    \psi'(7)=\{N,Pyrrole/5\}\\
    \psi'(2)=\psi'(3)=\{C,Benzene/6,Pyrrole/5\}
\end{align*}
The edge-labelling
and neighborhood functions do not change
after relation-enrichment.
\end{myexample}

\noindent
The vertex-enriched graph thus extends the vertex-labelling of
a graph $G$, with the vertex-labels from the hypergraph $H$
obtained from relations $R_1,\ldots,R_k$ defined on $G$. The
resulting graph can be used directly by the implementations
of GNNs described in the appendix. We note that
the process of vertex-enrichment is a simplification
of the full relational information available. For example,
in the example above, if an atom (represented
by a vertex in the molecular graph) is part of
more than 1 benzene ring, then its vertex-enrichment
will only contain a single entry for $Benzene/6$, indicating
that it is part of 1 or more benzene rings. 

\begin{definition}
{\bf (Vertex-Enriched GNN)} 
Let $G = (V,E,\sigma,\psi,\epsilon)$, and $Relabel$,
$Vec$ and $NN$ be as before. Then, a Vertex Enriched
GNN is $VEGNN(G) = NN(Vec(Relabel(VE(G,\mathcal{R}))))$.
\end{definition}

\subsection{Practical Considerations}
\label{sec:gnn_prac}

The GNN variants in this paper differ
in the $Relabel$ operation, based on the convolution procedure
employed. In this work, we employ the following different convolution
procedures:

\begin{enumerate}
    \item Localised approximation to spectral graph convolution~\cite{Kipf2017gcn}: This is a spectral method for graph convolution that uses convolutional aggregator. This is a simple and well-behaved layer-wise propagation rule for neural network models which operate directly on graphs.
    \item Multi-scale graph convolution~\cite{morris2019weisfeiler}: This convolution method can
        perform convolution operations using multiple-sized neighbourhoods (the
        authors call this ``higher order'' graph convolution).
    \item Graph convolution with attention~\cite{velickovic2018graph}: This is a
        spatial method of graph convolution that uses an ``attention'' mechanism, that
        estimates the importance of vertices in the neighbourhood of a vertex.
    \item Sample-and-aggregate graph convolution~\cite{hamilton2017inductive}: Here
        the convolution procedure samples from a distribution that is constructed
        from feature-vectors of vertices in the neighbourhood of a vertex.
    \item Graph convolution based on auto-regressive moving average~\cite{bianchi2019graph}: This is a        convolution method that employs a polynomial function of the feature-vectors in the
        neighbourhood of a vertex.
\end{enumerate}

\noindent
The $Relabel$ operation also includes a pooling step after each
convolution operation. Additional details are in Appendix
\ref{app:GNNs}. In all cases, we have used a fixed
vectorisation function $Vec$ that is based on a readout mechanism, and
$NN$ refers to a standard multi-layer perceptron (MLP).

\vspace{0.2cm}

We now elaborate on three practical issues arising from the
use of Vertex-Enriched GNNs:

\begin{enumerate}
    \item The vertex-enriched graphs we obtain allow us to use standard forms of GNNs
        (see Procedure~\ref{proc:veproc}).
        However, this comes with the limitation that we only change the vertex-labellings.
        A GNN defined directly on hypergraphs would have
        access to more information than the vertex-enriched
        GNN, since the former would retain the edge-labelling
        on hyperedges, and can have a richer definition of
        the neighbourhood function. Recently, there have
        been some proposals of GNNs for
        hypergraphs~\cite{feng2019hypergraph,jiang2019dynamic,yadati2019hypergcn}.
        It is possible that these forms of GNNs may perform
        better than Vertex-Enriched GNNs. We expect that the results
        in Section~\ref{sec:results} will act as baseline for such comparisons.
    \item Procedure~\ref{proc:veproc} requires identification
        of subgraphs of the original graph. That is: for every relation $R_i \in \mathcal{R}$,
        the corresponding hyperedge $H_i$ is a
        subset of vertices $\{v_1, \ldots, v_n\} \in V$,
        such that $(v_1,\ldots,v_n) \in R_i$. This 
        step requires the identification of all
        subsets of vertices of the graph
        constituting hyperedge as above. 
        For a graph $(V,E)$, this can,
        in the worst case require an
        examination of $\binom{|V|}{n}$ combinations.
        Therefore, for arbitrary sized graphs and
        subgraphs, this
        is computationally hard. In practice, we will be forced to
        impose bounds on the size of $V_s$ and on the
        size of the subgraph.
    \item We have not described how the relations in ${\cal R}$ themselves are
        obtained. There are two possibilities here. First, they
        are provided as prior information ({\em background knowledge} in
        ILP terminology). Secondly, the ${\cal R}$ provided as prior
        information can be augmented by relations constructed
        automatically (see Procedure~\ref{proc:augproc}).
        In this paper, the construction of new relations is
        done using an ILP engine, by adapting the usual clause-construction
        procedure (see Appendix~\ref{app:ilprel} for an ILP-based
        implementation of $LearnRels$ in Procedure~\ref{proc:augproc}).\footnote{
            Usually, clauses constructed by an ILP engine
            are either used as part of a hypothesis, or as features
            to construct a Boolean-vector representation of the data (``propositionalisation'').
            Here, the clauses are not used in either of these roles,
            but as relations that augment
            the prior knowledge available to the GNN.}
\end{enumerate}

\begin{algorithm}[!htb]
\SetAlgoLined
\KwData{Graph $G=(V,E,\sigma,\psi,\epsilon)$, a set of relations $\mathcal{R}=\{R_1, \ldots, R_k\}$}
\KwResult{Vertex-Enriched Graph, $G'=(V,E,\sigma,\psi',\epsilon)$}
 Let $\psi' := \psi$\;
 \For{$R_i \in \mathcal{R}$}{
  Let $R_i \subseteq V^n$\;
  ${\cal H}_i = \{\{v_1,\ldots,v_n\}: (v_1,\ldots,v_n) \in R_i\}$\;
  Let $V_s = \bigcup_{H_j \in {\cal H}_i}H_j$\;
  \For{$v_j \in V_s$}{
   $\psi'(v_j) := \psi'(v_j) \cup \{R_i/n\}$\;
  }
 }
 \Return $(V,E,\sigma,\psi',\epsilon)$\; 
 \caption{(\textbf{EnrichGraph}) Vertex-Enrichment of a graph $G$, given a
 set of relations $\mathcal{R}$. The new label of a vertex includes all
 the relations of which the vertex is part.}
 \label{proc:veproc}
\end{algorithm}

\begin{algorithm}[!htb]
\SetAlgoLined
\KwData{A graph $G = (V,E,\sigma,\psi,\epsilon)$;
    a set pre-classified instances $E = \{(G_i,y_i):$ 
         $G_i = (V_i,E_i,\sigma,\psi_i,\epsilon_i)~\mathrm{and}~y_i \in {\cal Y}\}$;
        a set of relations $\mathcal{R}=\{R_1, \ldots, R_k\}$; and
        a bound $n$ on the number of new relations}
\KwResult{A vertex-enriched graph $G'=(V,E,\sigma,\psi',\epsilon)$
    obtained from an augmentation of ${\cal R}$ by at most $n$ new relations
        obtained using $E$}
 Let ${\cal R}' = LearnRels({\cal R},E,n)$\;
 \Return $EnrichGraph(G,\mathcal{R} \cup \mathcal{R}')$\;
 \caption{(\textbf{AugmentRels}) Augmentation of a set of relations $\mathcal{R}$ by learning new relations from data.}
 \label{proc:augproc}
\end{algorithm}

\section{Empirical Evaluation}
\label{sec:expt}

\subsection{Aims}
\label{sec:aim}

Our aims in this paper is to investigate the incorporation of
background knowledge by GNNs. Specifically, using the term
Vertex-Enriched GNNs (VEGNNs) to denote the inclusion of relations into
GNNs (See Procedure \ref{proc:veproc}),
the experiments attempt to answer to the following questions: 

\begin{enumerate}
    \item How do VEGNNs perform against standard GNNs? This
        compares GNNs with and without the inclusion of
        domain-knowledge.
    \item Can the performance of VEGNNs be improved by using symbolic learner
        with access to the same domain-knowledge? This tests whether
        the computational machinery of a GNN is sufficient to 
        construct (representations of) the high-level relationships
       needed for good prediction. 
\end{enumerate}

\subsection{Materials}
\label{sec:mat}

\subsubsection{Data}
\label{sec:data}

The datasets are classification problems arising in the field of 
drug-discovery. We have evaluated our GNNs on 73 real-world 
binary classification datasets. Each dataset represents an extensive drug evaluation effort at the National Cancer Institute (NCI)\footnote{\url{https://www.cancer.gov/}}. The datasets represent experimentally determined effectiveness of anti-cancer activity of a compound against a number of cell lines~\cite{marx2003data}. The datasets correspond to the concentration parameter GI50, which is the concentration that results in 50\% growth inhibition. Some of the datasets have been used in various data mining studies such as in a study involving the use of graph kernels in machine learning~\cite{ralaivola2005graph}. 

\begin{figure}[h]
    \centering
    \begin{tabular}{|c|c|c|c|}
    \hline
    \# of Datasets & \begin{tabular}[c]{@{}c@{}}Avg. \# of Molecules per\\ dataset (Graphs)\end{tabular} & \begin{tabular}[c]{@{}c@{}}Avg. \# of Atoms per\\ molecule (Vertices)\end{tabular} & \begin{tabular}[c]{@{}c@{}}Avg. \# of Bonds per\\ molecule (Edges)\end{tabular} \\ \hline
    73 & 3032 & 24 & 51 \\ \hline
    \end{tabular}
    \caption{Summary of datasets (Total number of instances is 221306)}
\end{figure}

\subsubsection{Background Knowledge}
\label{sec:bk}

\begin{figure}[h]
    \centering
    \includegraphics[width=0.5\textwidth]{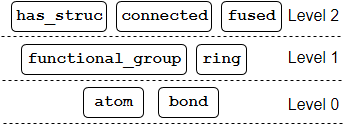}
    \caption{Levels of abstraction in the background knowledge~\cite{dash2018large}}
    \label{fig:back}
\end{figure}

\begin{figure}[!htb]
    \centering
    \includegraphics[width=0.8\textwidth]{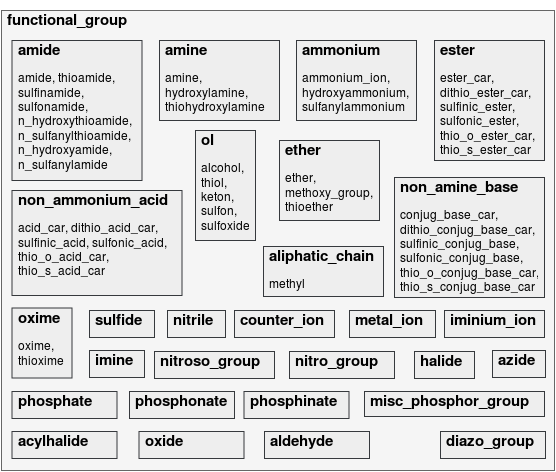}
    \caption{Functional group hierarchy}
    \label{fig:prophier}
\end{figure}

\begin{figure}[!htb]
    \centering
    \includegraphics[width=0.8\textwidth]{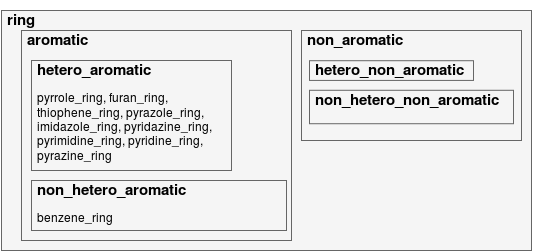}
    \caption{Ring hierarchy}
    \label{fig:ringhier}
\end{figure}

The initial version of the background knowledge in this paper here was used in \cite{Craenenbroeck2002dmax,ando2006discovering}. It is a collection of 
logic programs defining almost $100$ relations for various functional groups 
and ring structures in a chemical compound.\footnote{The definitions used
were originally developed for tackling industrial-strength problems 
by the biotechnology company PharmaDM.} The background
knowledge consists of multiple hierarchies.
However, we modified some of the predicate definitions to avoid redundant computation and for tractability to trade-off completeness for efficiency.
For proprietary reasons, we are only able to show the results of using the definitions, which are functional groups represented  as $\mathtt{functional\_group(CompoundID,Atoms,Length,Type)}$ and 
rings described as $\mathtt{ring(CompoundID,RingID,Atoms,Length,Type)}$. 
For efficiency, we have restricted the definition
of the ring relation to produce rings of maximum length 8.
The first use of this new version of the background knowledge is
reported in \cite{dash2018large} where we had also defined three
higher level relations to infer the presence of composite structures
from the presence of functional groups and rings in a compound.
These are: the presence of fused rings, connected rings and substructures.
These relations are defined below.
\begin{description}
    \item[$\mathtt{has\_struc(CompoundId,Atoms,Length,Struc)}$] This relation is $TRUE$ if a compound identified by $\mathtt{CompoundId}$
    contains a structure $\mathtt{Struc}$ of length $\mathtt{Length}$ containing a set of atoms in $\mathtt{Atoms}$.
    \item[$\mathtt{fused(CompoundId,Struc1,Atoms1,Struc2,Atoms2})$] This relation is $TRUE$ if a compound identified by $\mathtt{CompoundId}$
    contains a pair of fused structures $\mathtt{Struc1}$ and $\mathtt{Struc2}$ with
    $\mathtt{Atoms1}$ and $\mathtt{Atoms2}$ respectively (that is, there is
    at least 1 pair of common atoms).
    \item[$\mathtt{connected(CompoundId,Struc1,Atoms1,Struc2,Atoms2)}$] This relation is $TRUE$ if a compound identified by 
    $\mathtt{CompoundId}$
    contains a pair structures $\mathtt{Struc1}$ and $\mathtt{Struc2}$ that
    with
    $\mathtt{Atoms1}$ and $\mathtt{Atoms2}$ respectively that are not fused
    but connected by a bond between an atom in $\mathtt{Struc1}$ and an atom in
    $\mathtt{Struc2}$.
\end{description}
The level of abstraction in the
background knowledge is shown in Fig.~\ref{fig:back}.
The hierarchy available in functional groups and rings is shown in Fig.~\ref{fig:prophier} and Fig.~\ref{fig:ringhier}.

\subsubsection{Algorithms and Machines}

The data used for this work and the set of symbolic relations ($\mathcal{R}$) described in Section~\ref{sec:bk} are written
as Prolog facts. For generating the additional set of ILP relations ($\mathcal{R}'$), we use Aleph~\cite{srinivasan2001aleph}
that takes the data and the background-knowledge as input.
This additional set of relations $\mathcal{R}'$ further augments
the existing relations in $\mathcal{R}$ for our $VEGNN'$ studies. 
A logic program extracts a set of vertices in a graph 
for which any symbolic relation $R_i$ ($\in \mathcal{R}$ or $\in \mathcal{R}'$) is $TRUE$. We use YAP compiler for execution
of this logic program.

The GNN variants used here are described in Appendix~\ref{app:GNNs}.
All the experiments are conducted in Python environment.
The GNN models have been implemented by using the PyTorch Geometric library~\cite{fey2019fast}, which is a geometric deep learning extension for PyTorch~\cite{paszke2019pytorch} and it provides graph pre-processing routines and makes the definition of graph convolution easier to implement.

For all the experiments, we use a machine with Ubuntu (16.04 LTS) operating system, and hardware configuration such as:
64GB of main memory, 16-core Intel Xeon processor, a NVIDIA P4000 graphics processor with 8GB of video memory.

\subsection{Method}
\label{sec:method}

In all experiments, we refer to GNN variants
as ${GNN}_{1,\ldots,5}$. The corresponding vertex-enriched
versions are ${VEGNN}_{1,\ldots,5}$.
The GNN variants have 1 hyper-parameter that determines the
structure of the GNN (see Appendix \ref{app:GNNs}). We will
denote this by $m$ and assume that it takes values from
a fixed-set of values $M$.
\\

\noindent
\textbf{Experiment 1: GNNs vs. VEGNNs}
\vspace{0.3cm}

For constructing the VEGNNs, we assume that we 
have access to a set of domain relations $\mathcal{R}$. The method used is as follows.

\begin{itemize}
    \item[] For each dataset $D$:
    \begin{enumerate}
        \item Let $Tr, Val, Te$ denote a train-validation-test
            split of the data $D$
        \item For each of ${GNN}_{1,\ldots,5}$ and ${VEGNN}_{1,\ldots,5}$:
        \begin{enumerate}
            \item Find the best value $m^* \in M$ using
                the performance on $Tr$ and $Val$
            \item Record the predictive performance on $Te$ of the
                model constructed using $m^*$
        \end{enumerate}
        \item Compare the
            performance of ${GNN}_i$ against that of ${VEGNN}_i$ ($i = 1,\ldots,5$).
        \end{enumerate}
\end{itemize}

\noindent
The following additional details are relevant:
\begin{itemize}
    \item The relations in ${\cal R}$ are those
        described in Section \ref{sec:bk}.
    \item In our implementation, we use three graph convolution blocks and
    three pooling blocks interleaving each other. 
    \item The convolution
    blocks can be of one of the five convolution variants listed in Section~\ref{sec:gnn_prac}.
    Due to the large-scale experimentation (number of datasets, number of GNN variants), the various hyperparameters in convolution
    blocks are set to default values in PyTorch Geometric library.
    \item The graph pooling block uses self-attention pooling~\cite{lee2019self} with pooling ratio of 0.5. We use a hierarchical pooling architecture that uses the readout mechanism proposed by Cangea \textit{et al.}~\cite{cangea2018towards}. The readout block aggregates node features to produce a fixed size intermediate representation for 
    the graph. The final fixed-size representation
    for the graph is
    obtained by element-wise addition ($\oplus$) of the three readout representations.
    \item The final representation is then fed as input to a 3-layered MLP. We use a dropout layer with fixed dropout rate of 0.5 after first layer of MLP.
    The loss function is negative log-likelihood between the targets and the predictions 
    from the model.
    Further detail on the GNN architectures is provided in Appendix~\ref{app:strGNN}.
    \item We select amongst two possible values of the
    structure hyperparameter $m$ (8 and 128),
        corresponding to small and large amounts of convolution in
        the convolutional-layers of the GNNs and VEGNNs;
    \item We use Adam~\cite{kingma2014adam} optimiser for training the GNNs ($GNN_{1,\ldots,5}$) and VEGNNs ($VEGNN_{1,\ldots,5}$). The learning rate is 0.0005, weight decay parameter is 0.0001, momentum factors are the default values of $\beta_{1,2}=(0.9,0.999)$.
    \item Maximum number of training epochs is 1000. 
    The batch size is 128. 
    \item We use an early-stopping mechanism~\cite{prechelt1998early} to obtain the optimal model after training that can be used for evaluation on $Te$. The patience period for early stopping is 50.
    \item Comparison of performance is done using the Wilcoxon
        signed-rank test, using the standard implementation
        within MATLAB (R218b).
\end{itemize}

\vspace{0.3cm}

\noindent
\textbf{Experiment 2: VEGNNs with ILP-constructed Relations}
\vspace{0.3cm}

Given a set of generic relations ${\cal R}$, and some data, a VEGNN should,
in principle, be able to construct new (domain-specific) relations across its internal layers.
That is,  it may not be necessary to provide a VEGNN with anything more
than ${\cal R}$. In this experiment, we investigate the extent to which this holds
in practice, by evaluating the effects of augmenting ${\cal R}$ with higher-level
relations learned by ILP. The ILP procedure used to obtain these relations has
been described elsewhere (see Procedure~\ref{proc:augproc}). Our method is as
follows.

\begin{itemize}
    \item[] For each dataset $D$:
    \begin{enumerate}
        \item Let $Tr, Val, Te$ denote the train-validation-test
            split of the data $D$
        \item Let ${\cal R'}$ denote a set of new relations obtained
            using an ILP engine with access to ${\cal R}$ and $Tr \cup Val$
        \item Let ${VEGNN}_{1,\ldots,5}$ denote the VEGNNs obtained with
            ${\cal R}$ and ${VEGNN}_{1,\ldots,5}^\prime$ denote the VEGNNs
            with ${\cal R} \cup {\cal R'}$.
            
            For each of
            ${VEGNN}_{1,\ldots,5}$ and ${VEGNN}_{1,\ldots,5}^\prime$:
                \begin{enumerate}
                    \item Find the best value $m^*$ for the structure
                        hyperparameter $m$, using $Tr$ and $Val$
                    \item Record the predictive performance on $Te$ of the
                        model constructed using $m^*$
                \end{enumerate}
        \item Compare the
            performance of ${VEGNN}_i$ against that of ${VEGNN}_i^\prime$ ($i = 1,\ldots,5$).
        \end{enumerate}
\end{itemize}

\noindent
The following additional details are relevant:

\begin{itemize}
    \item The relations in ${\cal R}$ are those described
        in Section \ref{sec:mat}. 
    \item The construction of the $VEGNN$s is as in Experiment 1.
    \item The relations in ${\cal R}'$ are obtained using the ILP engine Aleph \cite{srinivasan2001aleph} with hide-and-seek sampling~\cite{dash2019discrete}.
    \item We repeat the comparisons for $|{\cal R}'| = 100$, $|{\cal R}'| = 500$, and $|{\cal R}'| = 1000$.
    \item ILP-constructed relations can be complex, and involve several vertices.
        To ensure tractability, we restrict the computation to detecting
        a single hyperedge (and not all hyperedges) corresponding to the
        ILP-constructed relation. This results in a loss of information.
   \item As in Experiment 1, comparisons will be in the form
    of a Wilcoxon signed-rank test, implemented within MATLAB (R2018b).
\end{itemize}

\subsection{Results}
\label{sec:results}

\begin{figure}[!htb]
\centering
\begin{subfigure}{.5\textwidth}
  \centering
  \includegraphics[width=\textwidth]{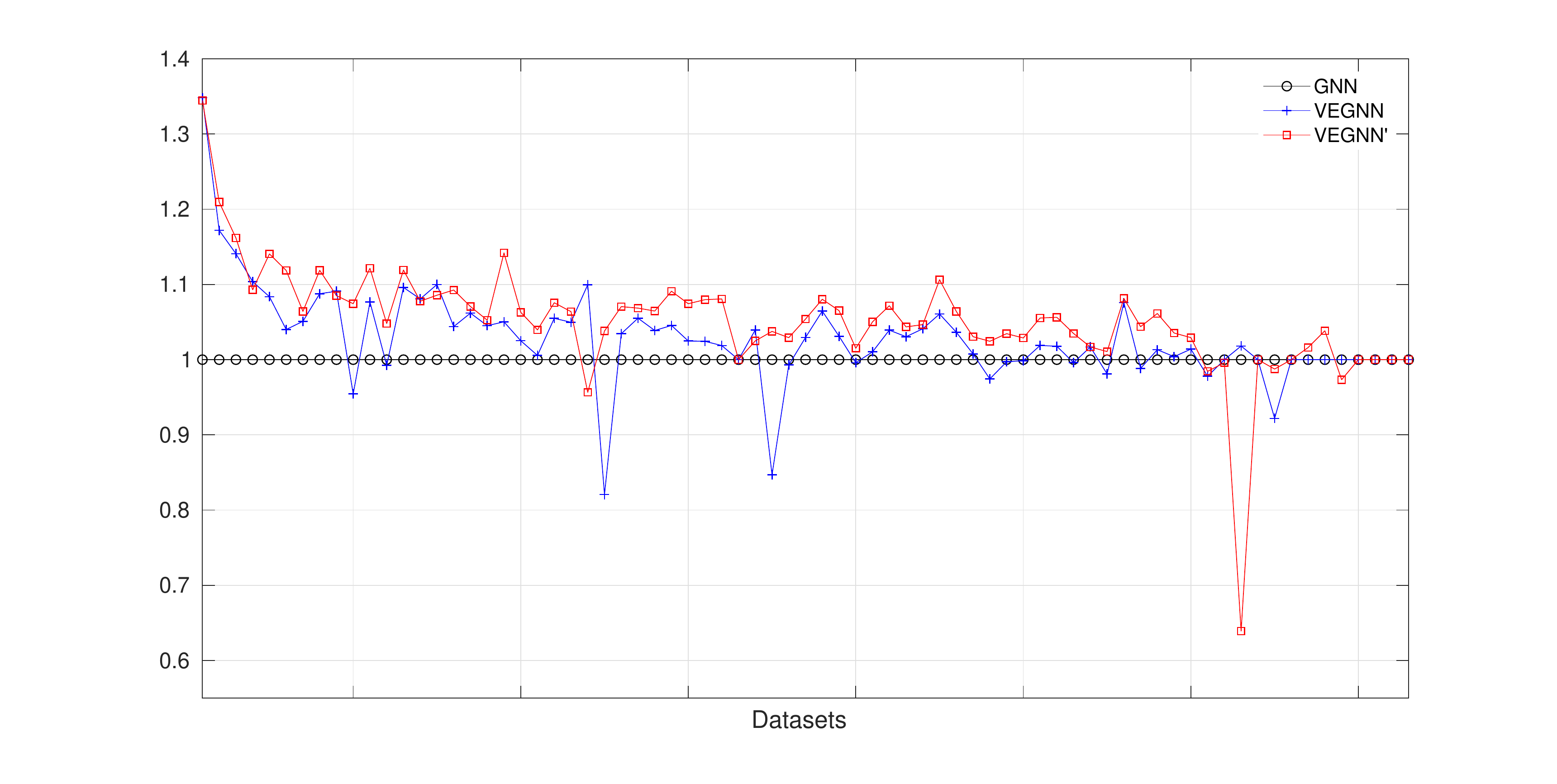}
  \caption{GNN variant: ${GNN}_1$}
\end{subfigure}%
\begin{subfigure}{.5\textwidth}
  \centering
  \includegraphics[width=\textwidth]{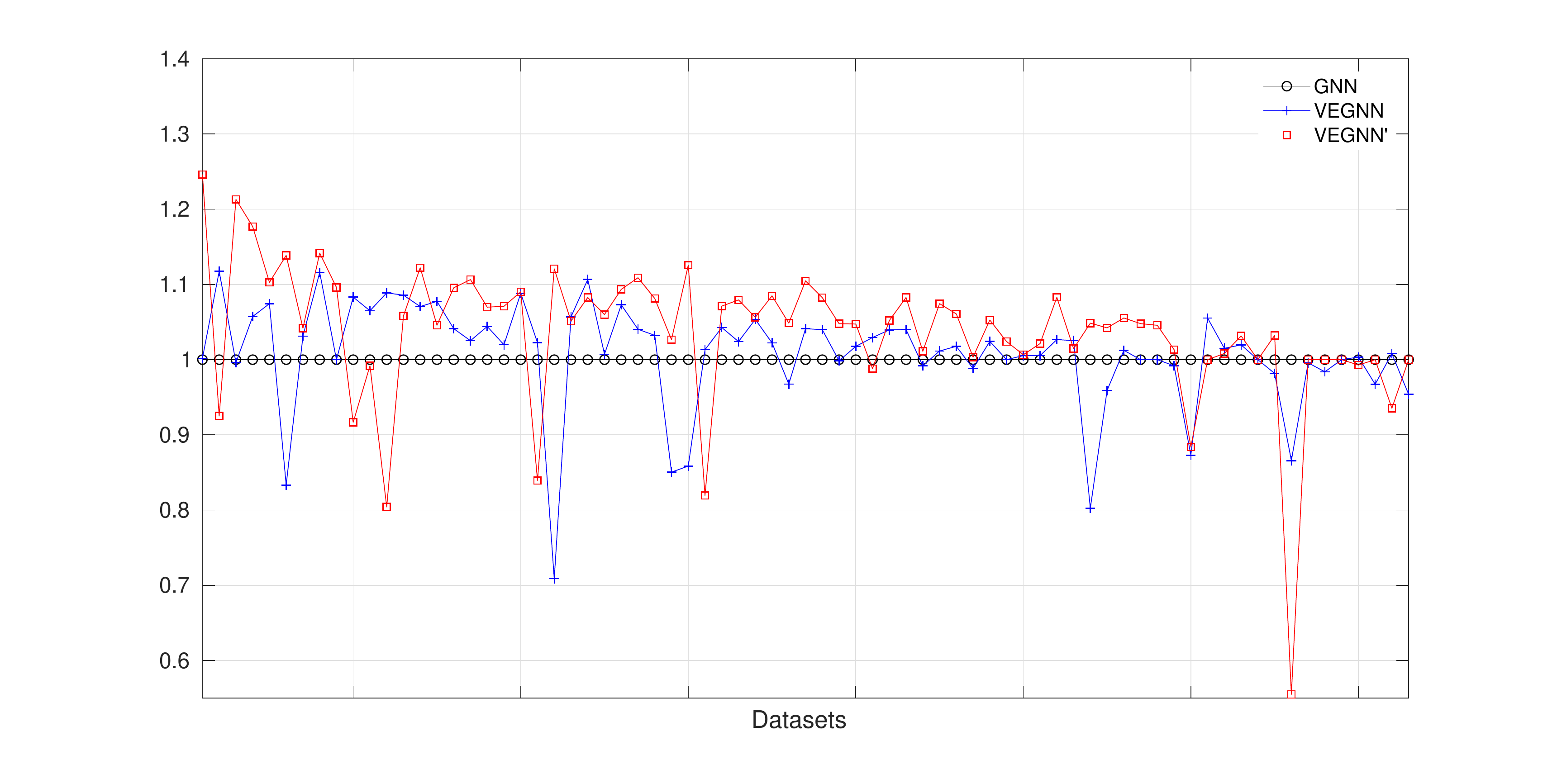}
  \caption{GNN variant: ${GNN}_2$}
\end{subfigure}
\begin{subfigure}{.5\textwidth}
  \centering
  \includegraphics[width=\textwidth]{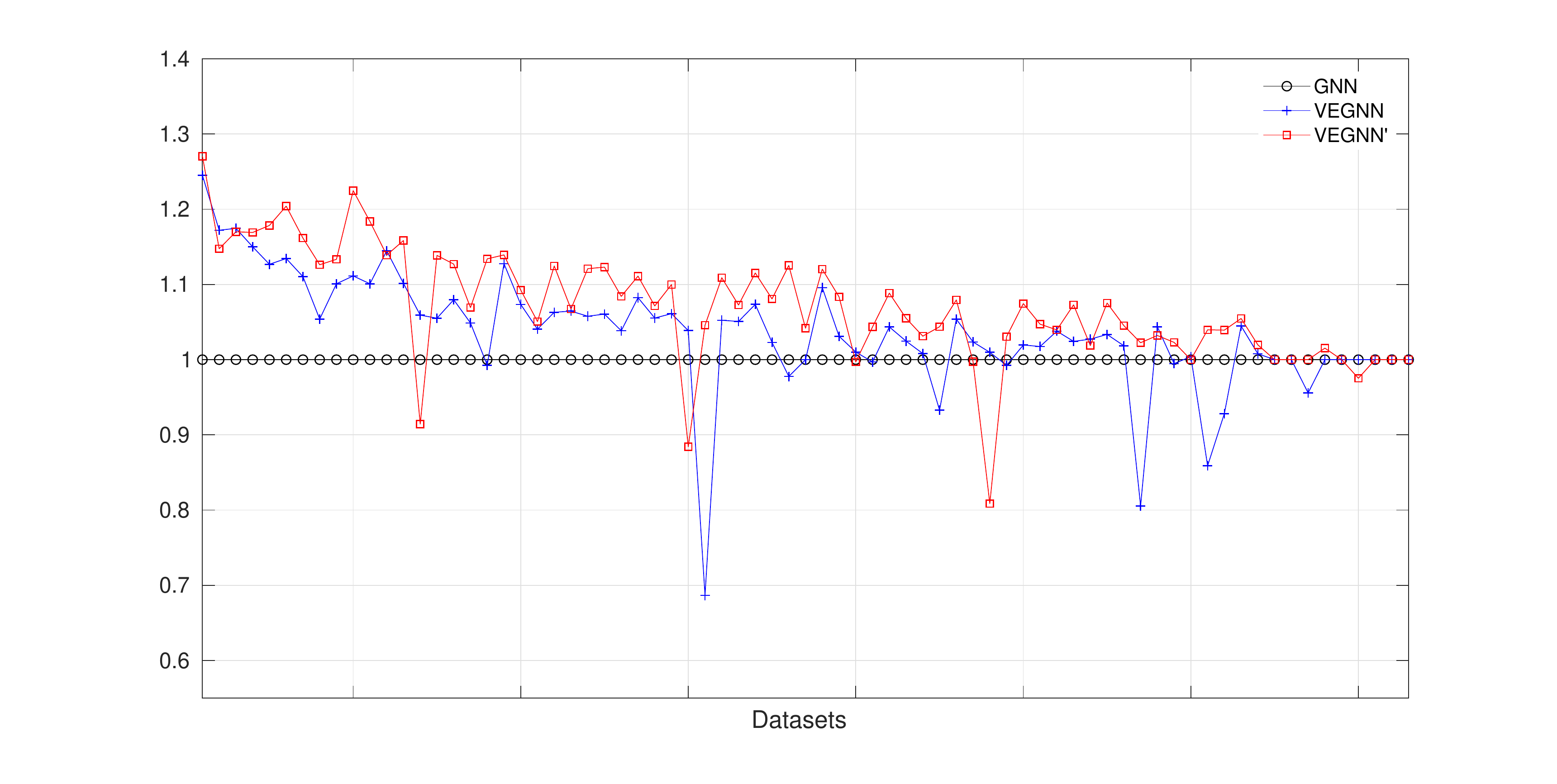}
  \caption{GNN variant: ${GNN}_3$}
\end{subfigure}%
\begin{subfigure}{.5\textwidth}
  \centering
  \includegraphics[width=\textwidth]{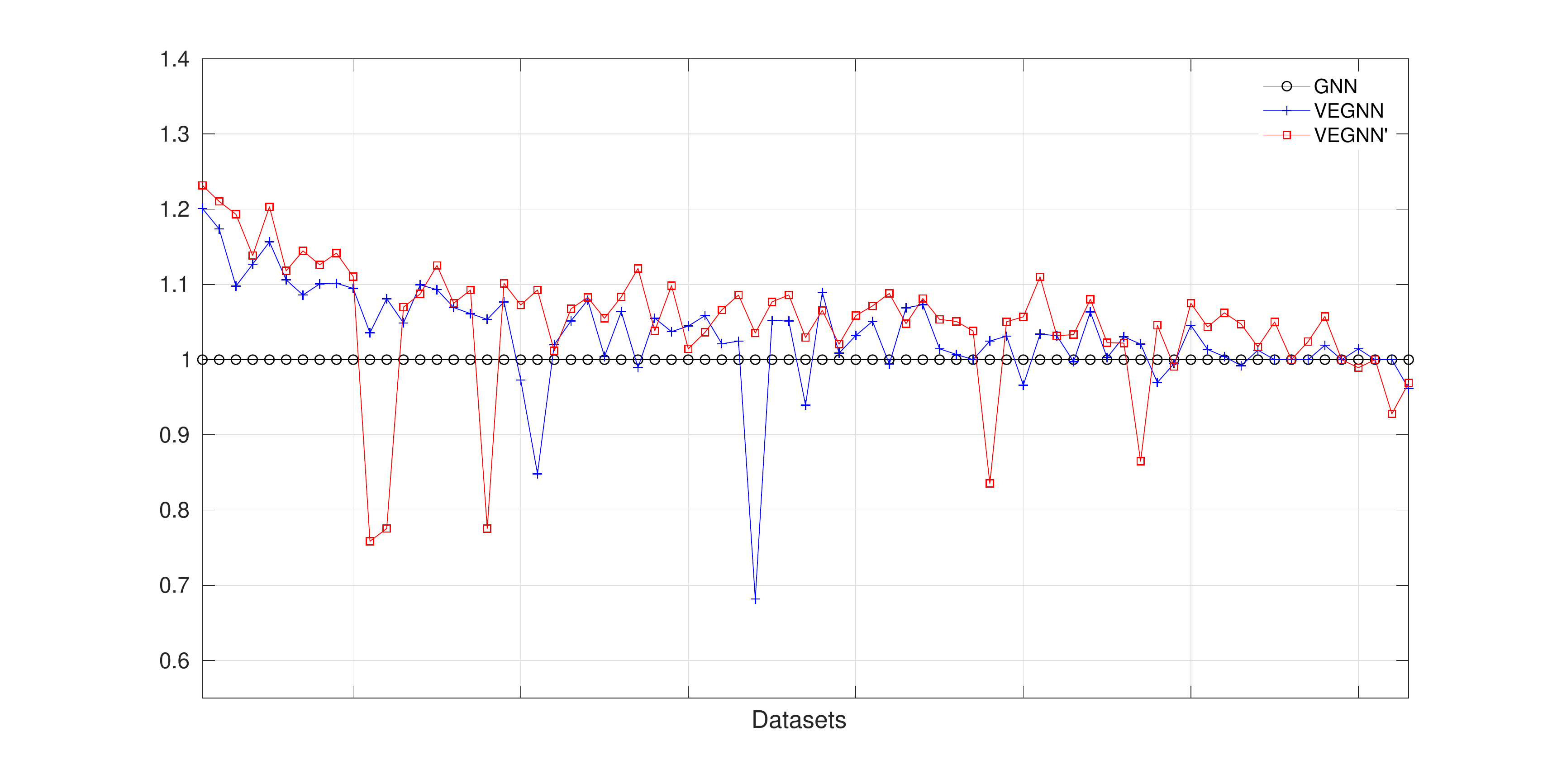}
  \caption{GNN variant: ${GNN}_4$}
\end{subfigure}
\begin{subfigure}{.5\textwidth}
  \centering
  \includegraphics[width=\textwidth]{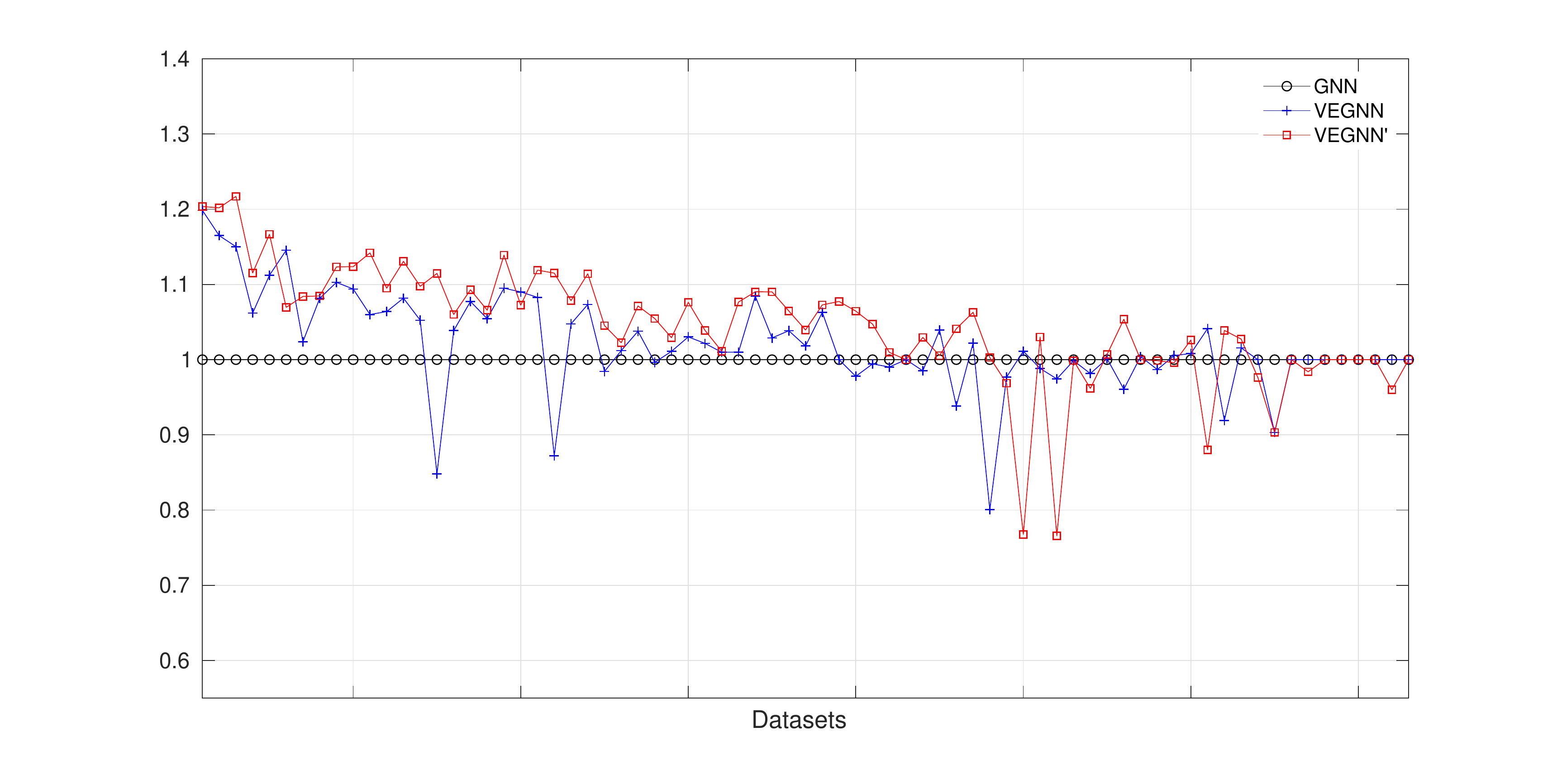}
  \caption{GNN variant: ${GNN}_5$}
\end{subfigure}
\caption{Qualitative comparison of graph-based neural networks.
Here GNN refers to the performance of the graph-based neural network without domain
relations; VEGNN refers to the performance of the network vertex-enriched with
generic domain relations
shown in Section \ref{sec:bk}; and VEGNN$'$ refers to the performance
of the network vertex-enriched with the generic domain-relations
and domain-specific relations constructed by an ILP
engine. Performance refers to predictive (holdout-set) accuracy, and
all performances are normalised against that of the GNN. Further, the compounds are arranged in order
of increasing GNN performance: the apparent trend of high-to-low gains for VEGNN and VEGNN$'$
from left to right are artifacts of this ordering. No significance should also be
attached to the line joining the data points: this is only for visual clarity.}
\label{fig:results1}
\end{figure}

The main results from the experiments are shown qualitatively
in Fig.~\ref{fig:results1}. The principal findings from the tabulations are these:
(a) Inclusion of domain-knowledge into GNNs (that is, the use of vertex-enriched
    GNNs) results in an improvement in predictive accuracy for all
    variants of GNN; and
(b) The performance of vertex-enriched GNNs can be improved further by augmenting
    the domain-relations with additional relations constructed by an ILP
    engine.

We now examine the results in more detail: From Fig.~\ref{fig:results1}, it
is evident that the performance of
graph-based networks improves with the inclusion
of domain-knowledge. A quantitative tabulation of
wins, losses and draws is in Fig.~\ref{fig:results2}.
These results provide
sufficient grounds to answer positively the primary research question addressed
in this paper, namely: do GNNs benefit from the inclusion
of domain-knowledge?

\begin{figure}[!htb]
\centering
\begin{tabular}{|c|l|} \hline
GNN & \multicolumn{1}{c|}{Accuracy ($VEGNN$ vs. $GNN$)}\\ 
Variant & \multicolumn{1}{c|}{Higher/Lower/Equal ($p$-value)} \\ 
\hline
${GNN}_1$ & 48/14/11 ($<$ 0.001) \\
${GNN}_2$ & 48/19/6	(0.005) \\
${GNN}_3$ & 53/11/9	($<$ 0.001) \\
${GNN}_4$ & 54/12/7	($<$ 0.001) \\
${GNN}_5$ & 43/19/11 (0.002) \\
\hline
\end{tabular}
\caption{Quantitative comparison of GNN performance. Here $GNN$ refers to the
graph-based neural network without domain-knowledge, and $VEGNN$ refers to the network
vertex-enriched with the generic domain-knowledge described in Section \ref{sec:bk}.
The tabulations are the number of
datasets on which $VEGNN$ has higher, lower or equal predictive accuracy on
a holdout-set. Statistical significance is assessed by the Wilcoxon
signed-rank test.}
\label{fig:results2}
\end{figure}

Assuming that it is useful to provide a GNN with domain-knowledge,
we can then ask: are vertex-enriched GNNs sufficiently powerful
to compute automatically any additional information
needed for high predictive performance?
The results in Fig.~\ref{fig:results1} suggest that the answer to this
is ``no'', since it appears that the inclusion of ILP-constructed relations makes 
a significant difference. To understand this better, we tabulate
quantitative differences obtained as the number of ILP relations
added is increased. This is shown in Fig.~\ref{fig:results3}.
The plot in Fig.~\ref{fig:results1}
uses 1000 ILP-relations (the corresponding quantitative
differences are the last column in Fig.~\ref{fig:results3}).

\begin{figure}[!htb]
\centering
\begin{tabular}{|c|l|l|l|}
\hline
 & \multicolumn{3}{c|}{Accuracy ($VEGNN'$ vs. $VEGNN$)} \\
GNN & \multicolumn{3}{c|}{Higher/Lower/Equal ($p$-value)} \\ \cline{2-4}
Variant & \multicolumn{1}{c|}{$|\mathcal{R}'| = 100$} & \multicolumn{1}{c|}{$|\mathcal{R}'| = 500$} & \multicolumn{1}{c|}{$|\mathcal{R}'| = 1000$} \\ \hline
${GNN}_1$ & 45/17/11 ($<$ 0.001) & 46/19/8 ($<$ 0.001) & 55/10/8 ($<$ 0.001) \\ 
${GNN}_2$ & 46/20/7 ($<$ 0.001) & 55/13/5 ($<$ 0.001) & 54/17/2	($<$ 0.001) \\ 
${GNN}_3$ & 47/17/9 ($<$ 0.001) & 49/16/8 ($<$ 0.001) & 55/12/6 ($<$ 0.001) \\ 
${GNN}_4$ & 40/27/6 (0.055) & 46/23/4 (0.013) & 53/16/4	($<$ 0.001) \\ 
${GNN}_5$ & 39/20/14 (0.026) & 49/14/10 ($<$ 0.001) & 51/13/9 ($<$ 0.001) \\ \hline
\end{tabular}
\caption{Quantitative comparison of performance after augmenting
    domain-relations with ILP-constructed relations. Here
    $VEGNN'$ denotes the vertex-enriched GNN obtained after
    augmenting the generic domain relations (${\cal R}$) with domain-specific
    relations constructed by an ILP engine (${\cal R}'$); and $VEGNN$
    denotes the vertex-enriched GNN with ${\cal R}$. The tabulations are the number of
    datasets on which $VEGNN'$ has higher, lower or equal predictive accuracy on
    a holdout-set. Statistical significance is assessed by the Wilcoxon signed-rank test.}
\label{fig:results3}
\end{figure}

Since the inclusion of even small numbers of ILP relations (100)
seems to improve performance of the VEGNN, it would appear that the internal
representations within a VEGNN are of limited expressivity
when compared to those constructed by ILP. In turn, the complete tabulation
suggests that a hybrid VEGNN-ILP learner is very likely to
be better than just a VEGNN learner (and in turn, a GNN learner).

We note that vertex-enrichment is only a vertex-related operation.
It is relevant to ask if there are any edge-related operations
associated with the  addition of domain-relations. Since these
relations result in hyperedges, a natural edge-operation is one
of {\em clique-expansion\/}~\cite{zhou2007learning} of the domain-relations. That is, the original graph is
transformed to a new graph by the inclusion of all pairwise edges
between vertices in hyperedges entailed by the relations. We have investigated this,
but for reasons of space, do not include the results here. A
summary of the effect of clique-expansion is:
(a) By itself, clique-expansion of domain-relations is not helpful;
(b) Clique-expansion, in combination with vertex-enrichment does not yield any clear
advantage over vertex-enrichment alone across the
GNN variants.


\section{Related Work}
\label{sec:relwork}

GNN-like models were first proposed in~\cite{sperduti1997supervised,baskin1997neural}.
In these studies, the features from the graph data was extracted using neural networks. 
Gori \textit{et al.}~\cite{gori2005new} and Scarselli \textit{et al.}~\cite{scarselli2008graph}
proposed new graph-based learning methods that used recursive aggregation
of information. They called these models `graph neural networks (GNNs)'.
The major boost to the field of GNNs followed the introduction of graph convolution~\cite{Kipf2017gcn} 
and the notion of graph embedding~\cite{cui2018survey,zhang2018network}.
Many such embedding methods are based on
iterative processing of the neighborhood information 
of any vertex. One such vertex embedding method was
formulated by generalising the convolution operation
to graphs. The convolution operation computes ``hidden''
states (essentially vector-representations) of the vertices in the graph.
There are a wide variety of convolution-based GNNs most
of which are classified into spectral- or non-spectral
(spatial) approaches. Two methodical and comprehensive surveys over a series of variants of graph neural networks can be found in \cite{zhou2018graph} and \cite{wu2020comprehensive}. We have already
seen that for practical problems the 
data cannot effectively be modelled by pairwise
associations. Methods have been
proposed to define convolutions for higher-order graphs or
hypergraphs~\cite{feng2019hypergraph,jiang2019dynamic,yadati2019hypergcn},
although none of these have considered the problem of inclusion of 
domain-knowledge. To the extent that we consider a vertex-enriched
graph to be a result of a hypergraph representation of the data,
the work proposed in this paper loosely falls under the category of Hypergraph-based
neural networks. 

Notwithstanding the convolution operation used in GNNs, one
drawback that has been identified is that the representations learned
by them could be poor if the amount of training data (number
of graphs) is small, which would lead to
poor generalisation~\cite{xu2018how}. The usual solution
to this problem is to overcome data scarcity by the use of
prior knowledge, a feature that is at the heart of
Inductive Logic Programming. In almost all applications of
ILP to date, the use of prior or {\em background knowledge}
is central (see \cite{muggleton2012ilp}).
In contrast, the position taken in the neural-network
literature, especially those dealing with networks with large
numbers of hidden layers, is that provided sufficient data are
available, representations of relevant domain-concepts can be
computed automatically from data. But when data are scarce,
this assumption breaks down. The
area of neuro-symbolic modelling~\cite{besold2017neural}
has been concerned with ways of combining symbolic and neural learning. A simple way of doing this has been
studied under the category of ``propositionalisation''
in ILP~\cite{lavravc1991learning,kramer2001propositionalization,krogel2003comparative,francca2014fast,francca2015neural}.
Although, propositionalisation approaches have been successfully applied to various problems but are still considered as ad hoc approaches. These approaches are 
studied in the larger context of macro-operators~\cite{castillo2002macro}, which
are approaches to improve the heuristic search in ILP systems and extract higher-level or meta-rules~\cite{alphonse2004macro}. 
Pioneering work on the combination
of neural-networks and symbolic features has been done by d'Avila Garcez and Zaverucha~\cite{gerson:c2ilp} 
and extended in Fran{\c{c}}a et al.~\cite{francca2014fast,gerson:rulext}. 
There are several studies that report that the relational features constructed using propositionalisation-based
approach can substantially improve predictive performance of statistical machine learning models, see for example:~\cite{ramakrishnan2007using,saha2012kinds}. 
Recently, ILP-based feature-construction for
deep multi-layer perceptrons (a special case of
Deep Relational Machines, or DRMs
\cite{lodhi2013deep})
was shown to yield
surprisingly good results on the datasets used here,
albeit with very large numbers of features \cite{dash2018large,dash2019discrete}. At the other
end of the spectrum, methods are now being developed
that include ``neural'' predicates (predicates whose
definitions are implemented by neural networks)
as part of the background knowledge available
to a symbolic learner \cite{de2019neuro}.

Domain-knowledge is often available as knowledge graphs (or semantic networks) rather than as 
a set of relations defined in logic.
Knowledge graph embedding~\cite{ding2018improving,ziegler2017injecting}
is a technique that is mostly applied to construct a vector representation
for the knowledge graph, which can then be
{\em infused} 
into  some form into a neural network. In
recent reports, it is proposed that the 
latent representation
learned by a neural network can be coupled with
the representation of the knowledge graph that
may improve the predictive performance of the
neural network model~\cite{gaur2019shades,kursuncu2019knowledge}.

\section{Conclusions}
\label{sec:concl}

Our focus in this paper has been on the use of graph-based 
neural networks (GNNs) on scientific data. Scientific understanding
is largely an incremental process that builds on knowledge that
is already known. It is natural therefore to expect that
automatic techniques intended for scientific data analysis
will similarly be able to utilise such knowledge. The results
here clearly show the benefit of having mechanisms to incorporate
domain-knowledge into GNNs. They also show the benefits of ILP
as a mechanism for identifying relationships that appear not
to be within the practical reach of the GNN variants we have
considered. An ILP-purist could well ask: why then should we use
GNNs at all? There are several reasons to persist, chief amongst
which are reasons of implementation efficiency and widespread
availability of packaged libraries. Assuming GNNs are useful,
our goal has been to show that they can be  more useful if
they use domain-specific relations, and yet more so if they include results
from an ILP engine.\footnote{The use of ILP would seem to undermine
the motivation just given for using GNNs. However, this is not so.
First, once the ILP relations are constructed, the main modelling
effort is still done using GNNs. Secondly, the construction of
relations is task that can be implemented by
a specialised library.}

To the best of our knowledge, the experiments in this paper constitute
some of the most extensive applications of GNNs to large-scale real-world scientific
data. It
has not been the focus of this paper to construct a GNN-based benchmark
for the data, but to investigate the use of domain-knowledge. There is undoubtedly
room in the future for comparative studies against other techniques that may or
may not utilise the domain-knowledge available. More immediately,
the process of vertex-enrichment can create very large vectors
at each vertex (the result of a many-hot encoding of the relations in
the vertex's label). We conjecture that this situation can be
improved by performing some dimensionality-reduction at each vertex.
A straightforward option is to include some form of auto-encoder at each
vertex, before re-labelling. Vertex-enriched GNNs can probably be significantly improved
by directly working with Hypergraph GNNs (HGNNs). In principle, HGNNs
will have more information (like hyperedge labels). Will HGNNs also benefit
from the use of ILP? We do not know the answer to this as yet.

Despite the recent empirical successes in various fields,
recent studies highlight some of the theoretical limitations
of GNNs. For instance, GNNs cannot distinguish between some pairs of graphs that are indistinguishable by the 1-WL test~\cite{xu2018how,morris2019weisfeiler}, that is, a GNN with any parameter setting cannot distinguish two graphs unless the labels of the graphs are same. 
A recent study on GNNs~\cite{Barcelo2020The} has shown that the class
of aggregate-combine GNNs cannot be logically more expressible than a fragment of 
two-variable first-order logic with counting quantifiers (Logic FOC2), which is a form of description logic.
In a different report, various theoretical limitations of
GNNs are studied, specifically, in terms of approximation
ratios of combinatorial algorithms~\cite{sato2020survey}.
We have already indicated that the vertex-enrichment procedure
described in this paper may not capture fully the relational
information present in the data. We believe
this limitation can be overcome by adopting a different
form of graph representation, that is nevertheless still
amenable to the use of GNNs. We intend to explore this as
future work.

At the outset of this paper, we motivated the use of machine learning
in developing an automated scientific assistant. While high predictive
power is expected from an ML-based scientific assistant, it is not sufficient.
It is evident that this paper's focus is on how prediction can improve
by the inclusion of domain-knowledge. An
Understandable explanation of the models constructed by GNNs remains a
challenge. 

\section*{Data and Code Availability}

Data, background-knowledge and codes used in our experiments are available at: \url{https://github.com/tirtharajdash/VEGNN}.

\section*{Acknowledgements}
This work is supported by DST-SERB Grant EMR/2016/002766, Government of India.
The second author is a Visiting Professorial Fellow at School of CSE, UNSW Sydney. We sincerely
thank Ing. Gustav {\v{S}}ourek, Czech Technical University, Prague for providing the dataset information; and researchers at the DTAI, University of Leuven, for suggestions on how to use the background knowledge within DMAX. We also thank Dr. Oghenejokpeme I. Orhobor and Professsor Ross D. King for providing us with
initial set of background-knowledge definitions.

\bibliography{main}

\appendix

\section{Graph Neural Networks}
\label{app:GNNs}

\subsection{Implementation}
\label{sec:gnn_impl}

In a graph $G=(V,E)$, let $X_v$ denote a vector that represents the labelling of a vertex $v \in V$. This is called the feature vector of the vertex $v$. In a GNN,
the $Relabel$ function is implemented by a neighbourhood aggregation mechanism\cite{xu2018how}. It updates the representation
of a vertex, $h_v$ iteratively. That is, in $k$th iteration (or $k$th layer), the representation of a vertex $v$, $h_v^{(k)}$ can be computed 
using two procedures: $\mathsf{AGGREGATE}$ and $\mathsf{COMBINE}$.
\begin{align}
    a_v^{(k)} &= \mathsf{AGGREGATE}^{(k)}\left(\left\{h_u^{(k-1)}: u \in \mathcal{N}(v)\right\}\right),\\
    h_v^{(k)} &= \mathsf{COMBINE}^{(k)}\left(h_v^{(k-1)},a_v^{(k)}\right)
\end{align}
where, $\mathcal{N}(v)$ denotes the set of vertices adjacent to $v$. Initially (at $k = 0$), $h_v^{(0)} = X_v$.

The $Vectorise$ function constructs a vector representation
of the entire graph. This step is
carried out after the representations of all
the vertices are relabelled by some iterations.
The vectorised representation of the entire graph can be obtained using a $\mathsf{READOUT}$ function that aggregates vertex 
features from the final iteration ($k = K$):
\begin{equation}
    h_G = \mathsf{READOUT}\left(\left\{h_v^{(K)} \mid v \in G \right\}\right)
\end{equation}

There are different variants of $\mathsf{AGGREGATE}$-$\mathsf{COMBINE}$ procedures available in
the literature on GNNs. These are mostly implemented using
the methods known as graph convolution and graph pooling (refer \cite{zhou2018graph,wu2020comprehensive}). The $\mathsf{READOUT}$ procedure is usually implemented using a global or hierarchical pooling operation. The convolution operations of various GNNs used in our work are briefly described in Appendix~\ref{app:convvariants}. Further, we use an additional pooling layer called structural-attention pooling after each of the convolution layer. This is briefly described in Appendix~\ref{app:graphpool}.

\subsection{Graph Convolutions}
\label{app:convvariants}

\subsubsection{Variant 1}
\label{app:gnnvariant1}
The first variant of GNN used in our work is based on spectral-based graph convolutional network proposed by Kipf and Welling ~\cite{Kipf2017gcn}. It uses a layer-wise (or iteration-wise) propagation rule for a graph with $N$ vertices as:
\begin{equation}
    \mathbf{H}^{(k)} = \sigma \left(\Tilde{D}^{-\frac{1}{2}}\Tilde{A}\Tilde{D}^{-\frac{1}{2}}\mathbf{H}^{(k-1)}\Theta^{(k-1)}\right)
    \label{eq:gcnconv}
\end{equation}
where, $H^{(k)} \in \mathbb{R}^{N \times D}$ denotes the matrix of vertex representations of length $D$, $\Tilde{A} = A + I$ is the adjacency matrix representing an undirected graph $G$ with added self-connections, $A \in \mathbb{R}^{N \times N}$ is the graph adjacency matrix, $I_N$ is the identity matrix, $\Tilde{D}_{ii} = \sum_{j}\Tilde{A}_{ij}$, and $\Theta^{(k-1)}$ is the iteration-specific trainable parameter matrix, $\sigma(\cdot)$ denotes the activation function e.g. $\mathrm{ReLU}(\cdot) = \max(0,\cdot)$, $\mathbf{H}^{(0)} = \mathbf{X}$, $\mathbf{X}$ is the matrix of vertex feature vectors $X_i$s.

\subsubsection{Variant 2}
\label{app:gnnvariant2}
The second variant is based on the graph neural network proposed by Morris \textit{et al.}~\cite{morris2019weisfeiler} that passes
messages directly between subgraph structures inside the graph. At iteration $k$, the feature representation of a vertex is computed by using
\begin{equation}
    h^{(k)}_{u} = \sigma \left( h^{(k-1)}_{u}\cdot \Theta^{(k)}_1 + \sum_{v \in \mathcal{N}(u)}{h^{(k-1)}_{v}\cdot \Theta^{(k)}_2} \right)
    \label{eq:graphconv}
\end{equation}
where, $\sigma$ is a non-linear transfer function applied component wise to the function argument, $\Theta$s are the layer-specific
learnable parameters of the network.

\subsubsection{Variant 3}
\label{app:gnnvariant3}
The third variant is an attention-based model, which is popularly known as Graph Attention Network (GAT)~\cite{velickovic2018graph}. This network assumes that the contributions of neighboring vertices to the central vertex are not pre-determined which is the case in the Graph Convolutional Network~\cite{Kipf2017gcn}. This adopts attention mechanisms to learn the relative weights between two connected vertices. The graph convolutional operation at iteration $k$ is thereby defined as:
\begin{equation}
    h^{(k)}_{u} = \sigma\left(\sum_{v \in \mathcal{N}(u) \cup u}{ \alpha_{uv}^{(k)}\Theta^{(k)}h^{(k-1)}_{u}} \right)
\end{equation}
where, $h^{(0)}_u = X_u$. The connective strength between the vertex $u$ and its neighbor vertex $v$ is called attention weight, which is defined as
\begin{equation}
    \alpha^{(k)}_{uv} = \mathrm{softmax}\left(\mathrm{LeakyReLU}\left( a^\mathsf{T}\left[\Theta^{(k)}h^{(k-1)}_u \mathbin{\|} \Theta^{(k)}h^{(k-1)}_v \right]\right)\right)
\end{equation}
where, $a$ is the set of learnable parameters of a single layer feed-forward neural network.

\subsubsection{Variant 4}
\label{app:gnnvariant4}
The fourth variant is called GraphSAGE and it is a framework for inductive representation learning on large graphs~\cite{hamilton2017inductive}. It is done in two steps: local neighborhood sampling and then aggregation of generating
the embeddings of the sampled nodes. GraphSAGE is used to generate low-dimensional vector representations for nodes, and is especially useful for graphs that have rich node attribute information.
The following is an iterative update of the node embedding:
\begin{equation}
    h^{(k)}_{u} = \sigma \left( h^{(k-1)}_{u}\cdot \Theta^{(k)}_1 + \frac{1}{|\mathcal{N}(u)|} \sum_{v \in \mathcal{N}(u)}{h^{(k-1)}_{v}\cdot \Theta^{(k)}_2} \right)
    \label{eq:sageconv}
\end{equation}
where, $\sigma$ is a non-linear transfer function applied component wise to the function argument, $\Theta$s are the layer-specific
learnable parameters of the network.

\subsubsection{Variant 5}
\label{app:gnnvariant5}
This variant of GNN is inspred by the auto-regressive moving avarage (ARMA) filters that are considered to be more robust than polynomial
filters~\cite{bianchi2019graph}. The ARMA graph convolutional operator
is defined as follows:
\begin{equation}
    \mathbf{H}^{(k)} = \frac{1}{M} \sum_{m=1}^{M}{\mathbf{H}^{(K)}_m}
\end{equation}
where, $M$ is the number of parallel stacks, $K$ is the number of layers; and $\mathbf{H}^{(K)}_m$ is recursively defined as
\begin{equation}
    \mathbf{H}^{(k+1)}_{m} = \sigma\left(\hat{L}\mathbf{H}^{(k)}_{m}\Theta^{(k)}_2 + \mathbf{H}^{(0)}\Theta^{(k)}_2 \right)
\end{equation}
where, $\hat{L} = I - L$ is the modified Laplacian. The $\Theta$ parameters are learnable parameters. 

\subsection{Graph Pooling}
\label{app:graphpool}

Graph pooling methods apply downsampling mechanisms to graphs.
In this work, we use a recently proposed graph pooling method
based on self-attention~\cite{lee2019self}.
It uses graph convolution defined in Eq.~\eqref{eq:gcnconv} to obtain a self-attention score as given in Eq.~\ref{eq:sagpool} with the trainable parameter replaced by $\Theta_{att} \in \mathbb{R}^{N\times 1}$, which is a set of trainable parameters in the pooling layer.
\begin{equation}
    Z = \sigma \left(\Tilde{D}^{-\frac{1}{2}}\Tilde{A}\Tilde{D}^{-\frac{1}{2}}\mathbf{X}\Theta_{att}\right)
    \label{eq:sagpool}
\end{equation}
Here, $\sigma(\cdot)$ is the activation function e.g. $\tanh$.

\subsection{Structure of the GNNs}
\label{app:strGNN}
\begin{figure}[!htb]
    \centering
    \includegraphics[width=0.4\textwidth]{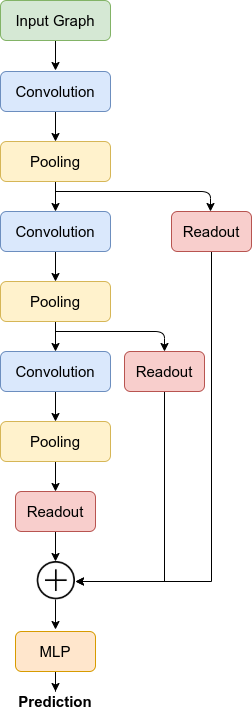}
    \caption{Graph classification architecture used in this work. We perform our experiments with five different types of graph convolution methods, each resulting in a different kind of GNN architecture.}
    \label{fig:schematic}
\end{figure}

The structure of the GNNs closely follows the structure
used in~\cite{lee2019self}. A schematic diagram of our implemented architecture is shown in Fig.~\ref{fig:schematic}.
As shown in the diagram, the output of the hierarchical pooling is fed as input to a multilayer perceptron (MLP). So, the 
input layer of the MLP contains $2m$ units, followed by two 
hidden layers with $m$ units and $\left \lfloor{m/2}\right \rfloor$ units respectively. The activation function used in the
hidden layers is $\mathtt{relu}$. The output layer size is
$|\mathcal{Y}|$ (in this work, 2) with $\mathtt{logsoftmax}$
activation.

\section{ILP Specifics}

\subsection{Extending Domain-Knowledge}
\label{app:ilprel}

We assume that a set of relations ${\cal R}$ are provided as part of the
background knowledge $B$ available to an ILP engine.\footnote{Besides ${\cal R}$, $B$ will
usually contain additional ILP-specific content like mode declarations (see 
\cite{mugg:progol}, along with search constraints and ancillary predicates).}
Given $B$ and data $E$ consisting of a set of positive and negative instances (here
representing molecules with or without the property of interest), and ILP engine
can construct new clauses defined in terms of the relations in ${\cal R}$. These
clauses can be additionally be ordered in terms of some utility function
(for example, a clause encoding a relation that holds for large number of
positive instances may have a high utility). The so-called technique of
ILP-based ``propositionalisation'', for example, identifies high-utility
clauses
(for example, see~\cite{ramakrishnan2007using,joshi2008feature,dash2018large}).
The procedure used to draw ``new'' relations using ILP-derived techniques
is in Procedure~\ref{proc:ilprel}.

\begin{algorithm}[h]
\SetAlgoLined
\KwData{Domain knowledge $B$, A set of examples $E$, and $MaxDraws$}
\KwResult{A set of relations $\mathcal{R}'$}
 $\mathcal{R}' := \emptyset$\;
 $draws := 0$\;
 $i := 1$\;
 $Drawn := \emptyset$\;
 \While{$draws \leq MaxDraws$}{
  Randomly draw an example $e_i \in E$ with replacement\;
  Let $\bot(B,e_i)$ be the most specific rule
    that entails $e_i$, given $B$\; \label{step:bot}
  Let ${\cal D}_C$ be a distribution over clauses\; \label{step:distr}
  Draw a clause $C_i$  using ${\cal D}_C$ s.t. $C_i \succeq_\theta \bot_d(B,e_i)$\; \label{step:drawclause}
  \If{$C_i$ is not redundant given $Drawn$}{
    \label{step:redun}
    Let $C_i = (Class(({\mathbf x},c) \leftarrow {Cp}_i({\mathbf x})))$\;
    Let $R_i = (NewR({\mathbf x}) \leftarrow {Cp}_i({\mathbf x}))$\;
    $Drawn ~:=~ Drawn \cup \{C_i\}$\;
    ${\cal R}' ~=~ {\cal R}' \cup \{R_i\}$\;
    increment $i$\;
    }
   increment $draws$\;
 }
 \Return $\mathcal{R}'$\; 
 \caption{(\textbf{LearnRels}) Procedure to construct new relations using ILP. We assume
the domain-knowledge consists of some relations ${\cal R} \in B$.
The construction of $\bot$
    is as
    described in~\cite{mugg:progol}, and $\succeq_\theta$ refers to Plotkin's $\theta$-subsumption~\cite{plotkin:thesis}.
    The redundancy test used is subsumption-equivalence.
    The distribution ${\cal D}_C$ is deliberately left unspecified here. In
    the experiments in the paper, ${\cal D}_C$ is either uniform (resulting in
    simple random construction of new relations), or a non-uniform selection
    based on clause-utility, as described in~\cite{dash2019discrete}. $NewR$
    is a new relation name that does not occur in $B$; $Cp$ denotes a conjunction
    of literals; ${\mathbf x}$ is
    shorthand for $x_1,x_2,\ldots,x_n$, the variables in $Cp$.}
 \label{proc:ilprel}
\end{algorithm}

\subsection{Input}

We use the ILP engine Aleph to construct the most-specific
rule above. Aleph requires the specification of a mode language, specifying the predicates
in ${\cal R}$. The mode-language used for the experiments
in the paper is given below:
\begin{verbatim}
    :- modeb(*,bond(+mol,-atomid,-atomid,#atomtype,#atomtype,#bondtype)).
    :- modeb(*,has_struc(+mol,-atomids,-length,#structype)).
    :- modeb(*,connected(+mol,#structype,-atomids,#structype,-atomids)).
    :- modeb(*,fused(+mol,#structype,-atomids,#structype,-atomids)).
\end{verbatim}
The `\#'-ed arguments in the mode declaration 
refers to type, that is, $\mathtt{\#atomtype}$ refers to the type of atom, $\mathtt{\#bondtype}$ refers to the type of bond, and $\mathtt{\#structype}$ refers to the 
type of the structure (functional group or ring) associated with the molecule. 

Each data instance (a molecule) is represented by a set of ground facts of
the following kind:
\begin{verbatim}
    bond(m1,27,24,o2,car,1).
    ...
\end{verbatim}
Here {\tt bond(m1,27,24,o2,car,1)} denotes that in instance {\tt m1}
there is an oxygen atom (id 27), and a carbon atom (id 24) connected
by a single bond ({\tt car} denotes a carbon atom in an aromatic ring).

Given the molecular structure additional facts like {\tt functional\_group/4} and {\tt ring/4} are pre-computed for efficiency using the generic relations in ${\cal R}$ (which contain the symbolic
definitions of benzene rings, oxide groups, {\em etc.}). This results in
facts like the following:
\begin{verbatim}
    functional_group(m1,[27],1,oxide).
    ring(m1,[25,28,30,29,26,23],6,benzene_ring).
    ...
\end{verbatim}

We note that these predicates result in
a {\em reification} of the predicates in ${\cal R}$
(that is, the predicate symbols are converted to terms).
The predicates $\mathtt{has\_struc}/4$, $\mathtt{connected}/5$ and $\mathtt{fused}/5$ are
defined over these predicates. For example (in Prolog format):

\begin{verbatim}
    has_struc(Mol,Atoms,Length,Type):-
        ring(Mol,Atoms,Length,Type).
    has_struc(Mol,Atoms,Length,Type):-
        functional_group(Mol,Atoms,Length,Type).
    ...
\end{verbatim}
We  reiterate that these predicates are  defined directly
on the relations in ${\cal R}$: the use
of  {\tt functional\_group/4} and {\tt ring/4}
is for compactness and efficiency.

\subsection{Output}

Given the mode language, and data consisting of the
molecular structure, the ILP engine finds clauses 
like these (shown as Prolog clauses):

\begin{verbatim}
    class(A,pos):-
        has_struc(A,D,E,ester_car),
        bond(A,F,G,c1,c1,3).
            
    class(A,pos):-
        connected(A,benzene_ring,D,benzene_ring,E),
        connected(A,keton,F,non_hetero_non_aromatic,G).
            
    class(A,pos):-
        fused(A,benzene_ring,D,imidazole_ring,E), 
        connected(A,oxide,F,oxide,G).
    ...
\end{verbatim}
\noindent
Each such clause is converted to an $n$-ary relation using
the steps in Procedure~\ref{proc:augproc}.

\end{document}